\documentclass{IEEEtran}

\usepackage{amssymb,amsmath,amsfonts}
\usepackage{algorithmic}
\usepackage{algorithm}
\usepackage{array}
\usepackage{booktabs}
\usepackage{colortbl}
\usepackage[hidelinks]{hyperref}
\usepackage{graphicx}
\usepackage{makecell}
\usepackage[mathlines, switch]{lineno}
\usepackage{multirow}
\usepackage[sort&compress, numbers]{natbib}
\usepackage{pifont}
\usepackage{stfloats}
\usepackage[caption=false,font=footnotesize]{subfig}
\usepackage{textcomp}
\usepackage{url}
\usepackage{verbatim}
\usepackage[dvipsnames]{xcolor}

\hypersetup{
  colorlinks = true,
  urlcolor = CadetBlue,
  linkcolor = Cerulean,
  citecolor = Maroon
}

\definecolor{light-gray}{rgb}{0.96, 0.96, 0.96}
\definecolor{blue}{rgb}{0.08, 0.33, 0.60}
\definecolor{purple}{rgb}{0.59,0.21,0.50}

\NewDocumentCommand{\codeword}{v}{%
\texttt{\textcolor{Black}{#1}}%
}

\begin{document}

\title{Tactics2D: A Highly Modular and Extensible Simulator for Driving Decision-making}

\author{Yueyuan~Li, Songan~Zhang, Mingyang~Jiang, Xingyuan~Chen, Yeqiang~Qian, Chunxiang~Wang, and~Ming~Yang 
\thanks{This work is supported in part by the National Natural Science Foundation of China under Grants 62173228 \textit{(Corresponding author: Ming Yang; e-mail: MingYANG@sjtu.edu.cn). }}
\thanks{Yueyuan Li, Mingyang Jiang, Yeqiang Qian, Chunxiang Wang, and Ming Yang are with the Department of Automation, Shanghai Jiao Tong University, Key Laboratory of System Control and Information Processing, Ministry of Education of China, Shanghai, 200240, CN. }
\thanks{Songan Zhang is with the Global Institute of Future Technology, Shanghai Jiao Tong University, Shanghai, 200240, CN.}
\thanks{Xingyuan Chen is with the University of Michigan-Shanghai Jiao Tong University Joint Institute, Shanghai Jiao Tong University, Shanghai, 200240, CN.}
}

\markboth{}%
{}

\IEEEpubid{}

\maketitle

\begin{abstract}

Simulation is a prospective method for generating diverse and realistic traffic scenarios to aid in the development of driving decision-making systems. However, existing simulators often fall short in diverse scenarios or interactive behavior models for traffic participants. This deficiency underscores the need for a flexible, reliable, user-friendly open-source simulator. Addressing this challenge, Tactics2D adopts a modular approach to traffic scenario construction, encompassing road elements, traffic regulations, behavior models, physics simulations for vehicles, and event detection mechanisms. By integrating numerous commonly utilized algorithms and configurations, Tactics2D empowers users to construct their driving scenarios effortlessly, just like assembling building blocks. Users can effectively evaluate the performance of driving decision-making models across various scenarios by leveraging both public datasets and user-collected real-world data. For access to the source code and community support, please visit the official GitHub page for Tactics2D at \url{https://github.com/WoodOxen/Tactics2D}.
\end{abstract}

\begin{IEEEkeywords}
  Simulation, Autonomous Vehicles, Testing, Decision-making.
\end{IEEEkeywords}

\section{Introduction}

Simulators boost the development of autonomous driving by saving time and resources. Particularly, in the case of driving decision-making, the simulators are expected to provide diverse traffic scenarios to help the models improve their capability to handle various real-world corner cases while avoiding safety issues. 

The ability to replay logs from various public trajectory datasets is crucial for obtaining diverse traffic scenarios. While CarRacing and highway-env are popular driving simulators acknowledged by the prominent reinforcement learning (RL) environment library, Gymnasium, they only offer specific kinds of pre-defined traffic scenarios \cite{carracing, highway-env}. A similar issue is observed with SMARTS, a multi-agent driving simulator \cite{zhou2020smarts}. This limited diversity makes them incapable of effectively evaluating driving decision-making methods intended to handle real-world challenges.

In addition to log replay functionality, the ability to customize maps is also in high demand. NuPlan and Waymax attempt to compensate by providing support for their own large-scale datasets \cite{caesar2021nuplan, gulino2023waymax}. However, these efforts have not been sufficient to address researchers' concerns about distribution biases and coverage of corner cases, which are common in log replay. As evidence of this, researchers continue to resort to building third-party toolkits for SUMO, a simulator primarily designed for microscopic traffic flow studies, due to its built-in support for map and trajectory editing \cite{sumorl, behrisch2011sumo}. Additionally, CommonRoad aimed to reform a driving simulator similar to SUMO with a data structure more suitable for decision-making \cite{althoff2017commonroad}. Unfortunately, due to inconsistencies within the development team, the modules of CommonRoad are not well integrated, leaving dissatisfaction in running efficiency.

Another challenge for driving simulation pertains to the realistic interaction between the agent vehicle and the NPCs\footnote{The term \textit{NPC}, commonly used in gaming, stands for non-player characters. In the context of driving simulation, it refers to simulator-controlled traffic participants, contrasting with vehicles controlled by the driving decision-making model, namely the agent.}. This issue has gained growing attention in recent years. InterSim has modeled the relationship between vehicles and resolved predicted conflicts \cite{sun2022intersim}. LimSim controls NPCs by predicting the time-to-collision (TTC), learning a distribution to randomize the risk the vehicles tend to take \cite{wenl2023limsim}. TBSim trained a bi-level imitation learning model to exploit realistic behavior from real-world driving logs \cite{xu2023bits}. However, a common limitation among these approaches is that the simulators are designed with a focus on specific methods. Since behavior modeling techniques are still evolving, building new simulators to accommodate these methods is inefficient.

User-friendliness is often overlooked in open-source driving simulators. The popular simulator CARLA demands significant storage and computation resources while lacking both forward and backward compatibility \cite{dosovitskiy2017carla}. Many other simulators \cite{carracing, althoff2017commonroad, highway-env, caesar2021nuplan, sun2022intersim, xu2023bits, wenl2023limsim, gulino2023waymax}, lack detailed documentation, making them less accessible for users seeking instant usability.

Tactics2D aims to offer a wide range of interactive traffic scenarios and extensive traffic components, facilitating a rapid and reliable development process for driving decision-making methods. The core features of Tactics2D include:

\begin{itemize}
    \item \textbf{Extensibility:} Tactics2D designs a highly modular and customizable code structure, allowing users to tailor methods and functions to their preferences. This encompasses road elements, traffic regulations, sensors, physics models, behavior models, traffic event detectors, and scenario evaluators.

    \item \textbf{Diversity:} Following the established paradigm, Tactics2D implements a range of essential methods and functions to ensure the diversity of traffic scenarios. This includes parsers for diverse open trajectory datasets, trajectory, and map formats, as well as pre-defined common road elements, behavior models, and map generators.

    \item \textbf{Usability:} Tactics2D emphasizes user accessibility, offering detailed and comprehensible tutorials and documentation. With cross-platform compatibility (Linux, MacOS, and Windows), the codebase of Tactics2D maintains high reliability, supported by over 85\% coverage of lines in the library through unit tests and integration tests.
\end{itemize}

\section{Design of Tactics2D}

\begin{figure*}[htb]
    \centering
    \includegraphics[width=\textwidth]{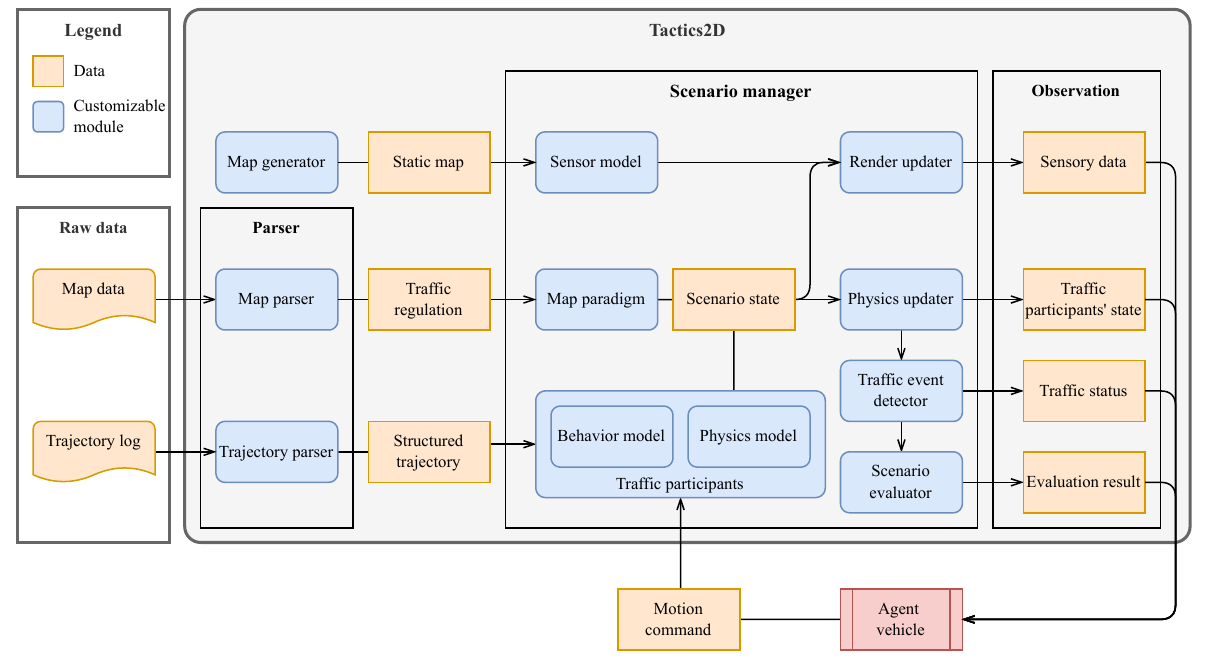}
    \caption{The structure of Tactics2D. The blue blocks are the executable code modules that the users can customize. The orange blocks are the data conveyed through the simulator.}
    \label{framework}
\end{figure*}

Figure \ref{framework} illustrates the functional modules in  Tactics2D. The following part of this section will explain each module's design rationale and details. It's worth noting that all code modules can be customized, provided they adhere to the prescribed paradigm. The complete documentation and tutorial are available at \url{https://tactics2d.readthedocs.io/en/latest/}.

\subsection{Parser}

The parser module standardizes map and trajectory data recorded in various formats into unified data structures.

\subsubsection{Map parser}

Within the map parser module, a list of robust parsers is employed to handle map formats under openly available standards, such as OpenDRIVE (XODR), OpenStreetMap (OSM), and Lanelet2-style OSM \citep{dupuis2010opendrive, haklay2008openstreetmap, poggenhans2018lanelet2}. XODR is commonly used to record high-resolution maps, while OSM has a mature platform to share global real-world map data. Lanelet2-style OSM is an improvement on the original, offering rich annotations that are frequently used. Thus, by supporting these formats, the module can effectively cover the diverse needs of users when importing custom maps.

\subsubsection{Trajectory parser}

 Tactics2D adopts a dataset-by-dataset approach to parsing trajectory data in different formats. This decision stems from the profound influence of open trajectory datasets on the paradigm by which the community records the trajectory data. Currently, this module supports the replay of a wide range of trajectory datasets, including the LevelX series, Argoverse, Dragon Lake Parking (DLP), INTERACTION, NuScenes, and Waymo Open Motion Dataset (WOMD) \citep{krajewski2018highd, bock2020ind, krajewski2020round, moers2022exid, shen2020parkpredict, zhan2019interaction, caesar2020nuscenes, ettinger2021large}. Each trajectory is instantiated as a traffic participant and forwarded to the scenario manager. If the dataset does not specify the type or size of the instance, a classifier trained using LevelX's data will match the traffic participant with internal templates based on its motion pattern.

For datasets lacking or prohibiting the public distribution of maps (the LevelX series, DLP, and INTERACTION),  Tactics2D offers manually annotated Lanelet2-style OSM maps. In cases where datasets combine maps and trajectories into a single file (NuScenes and WOMD), the map parsers are integrated into the dataset parsers.

\subsection{Map generator}

The map generator offers an alternative method to prepare maps for traffic scenarios besides the dataset replay mode. It generates maps either based on pre-defined rules or by extracting patterns from maps within datasets and recombining them into new configurations. Currently, the map generator can produce various types of maps, including parking lots of multiple difficulty levels, racing tracks that adhere to F1's requirements regarding length and curvature, highways with entries and exits, and unsignalized intersections.

\subsection{Scenario manager}

The scenario manager module acts as the central hub, facilitating information exchange and coordination among all other modules. It governs the creation and removal of structured maps, sensors, and traffic participants. It ensures the synchronization among updaters. Additionally, it decides when to allow the behavior controller to take over the NPCs and interact with the agent. Based on traffic status information from the traffic event detector, the scenario manager determines whether to terminate and reset the traffic scenario.

\subsubsection{Sensor model}

The sensor model defines how data should be rendered as an observation result. Two types of sensor information have already been provided in  Tactics2D: bird's eye view (BEV) semantic segmentation and single-line LiDAR point cloud. The BEV semantic segmentation result is expected to be output as RGB images, while the LiDAR point clouds are expected to be presented as 1D vectors.

\subsubsection{Map paradigm}

The map paradigm organizes road elements and traffic regulations while also defining their structures. Geometry information is crucial for static elements such as roads, areas, road lines, and nodes. Traffic regulations are associated with specific road elements. Each component of a map is assigned a unique identifier for easy access.

\subsection{Traffic participants in scenario manager}

All NPCs and agents are instantiated as traffic participants, each comprising essential components such as physics attributes (size, speed range, acceleration range, etc.), trajectory, behavior model, and physics model. In interactive mode, an NPC traffic participant receives motion commands from the behavior model, which are then used to update the physics state by the physics model. When an NPC traffic participant operates in log replay mode, it bypasses the behavior model and relies on the physics model to verify whether the physics state transition adheres to the physics constraints. 

\subsubsection{Behavior model}

The behavior model facilitates interaction between NPCs and the agent. Currently, the controller's behavior model is rule-based, covering functionalities such as emergency braking, car-following, and lane-changing triggered by factors like TTC, high speed, and short distance. A simplified version draws inspiration from theories similar to those employed in  SUMO \cite{lopez2018microscopic, erdmann2015sumo}. A more intricate collision avoidance mechanism is implemented based on methods utilized in Intersim \cite{sun2022intersim}. There are plans to augment the behavior models with learning-based approaches (refer to Section \ref{future}).

\subsubsection{Physics model}

Several dynamics or kinematics models are implemented in  Tactics2D. For pedestrians, a point mass model is recommended. The kinematics front-wheel driven bicycle model is suitable for bicycles, motorcycles, and low-speed vehicles, whereas the dynamics front-wheel driven bicycle model is preferable for high-speed vehicles. When a traffic participant replays a trajectory record, the physics updater verifies whether the trajectory record adheres to the physics models' constraints.

\subsection{Updaters in scenario manager}

\subsubsection{Render updater}

The render updater receives rendering rules from the sensor model and scenario state data from the map and traffic participant instances, converting them into sensory data. In off-screen mode, it outputs the data directly as matrices. In on-screen mode, it employs PyGame as the backend to render BEV semantic segmentation as RGB images and point clouds as grayscale images.

\subsubsection{Physics updater}

The physics updater is responsible for executing control commands within the system. It updates the physics state of traffic participants according to their backend physics models. The physics updater is designed to operate with a significantly smaller time step than the render updater, ensuring its precision. While the render updater is limited to updating at most 60 Hz (or every 100 ms), the time interval for the physics updater typically falls within the range of 5-10 ms. Despite this smaller time step, the physics updater's computational speed allows it to iterate 10 to 100 times faster, with the simulated time step being adjustable. In essence,  Tactics2D facilitates real-time simulation and supports fast-forwarding simulation, offering enhanced efficiency.

\subsubsection{Traffic event detector}

The traffic events detector is the module that monitors the status of the decision-making model-controlled vehicle. It continuously tracks two categories of issues: violation of traffic rules and within the simulation context. Violations of rules include collisions, retrograde movement, breaches of road lines, illegal turns, and disregard for red lights. Simulation exceptions cover prolonged inactivity, failure to reach the destination within a specified time frame, or driving beyond an assigned range. The traffic event detector runs at an editable frequency which cannot be higher than the physics updaters.

\subsubsection{Scenario evaluator}

The scenario evaluator is a customizable module specifically crafted for driving decision-making tasks. It can furnish step-by-step or episodic evaluation results based on the traffic status. Within Tactics2D, there are some pre-defined sets of evaluation criteria tailored for racing, parking, highway, intersection, and roundabout scenarios.

\section{Features of Tactics2D}

Table \ref{feature-comparison} offers a detailed comparison between Tactics2D and the influential\footnote{The identification of \textit{influential} in this paper is that the repository has over 100 citations counted by Google Scholar or over 100 stars on GitHub.} actively maintained\footnote{The identification of \textit{actively maintained} in this paper is that the simulator has any update within the last year.} open-source driving decision-making simulators, as referenced by \cite{li2024choose}. As evident from the table, the popular simulators have issues that undermine their functionality and efficiency, further detracting from the user experience.

\begin{table*}[htb]
    \centering
    \caption{Comparison of the key functionalities with the influential open-source simulators under active maintenance. \\ The simulators are listed in ascending order based on their release year.}
    \label{feature-comparison}
    \small
    \begin{tabular}{l | c c c c c c c} \toprule[2pt]
    \textbf{Simulator} & \parbox{1.6cm}{\centering\textbf{Built-in\\Environment}} & \parbox{1.6cm}{\centering\textbf{Custom\\Trajectory}} & \parbox{1.2cm}{\centering\textbf{Custom\\Map}} & \parbox{2cm}{\centering\textbf{Dataset\\Compatibility}} & \parbox{1.6cm}{\centering\textbf{Interactive\\NPCs}} & \textbf{Multi-agent} & \textbf{Lightweight} \\ \midrule
    SUMO \cite{behrisch2011sumo} & - & $\surd$ & $\surd$ & - & $\surd$ & - & $\surd$ \\
    \rowcolor{light-gray} CarRacing \cite{carracing} & $\surd$ & - & - & - & - & - & $\surd$ \\
    CARLA \cite{dosovitskiy2017carla} & $\surd$ & * & $\surd$ & - & $\surd$ & $\surd$ & - \\
    \rowcolor{light-gray} CommonRoad \cite{althoff2017commonroad} & $\surd$ & $\surd$ & $\surd$ & $\surd$ & $\surd$ & - & $\surd$ \\
    highway-env \cite{highway-env} & $\surd$ & - & - & - & $\surd$ & - & $\surd$ \\
    \rowcolor{light-gray} SMARTS \cite{zhou2020smarts} & $\surd$ & - & - & $\surd$ & $\surd$ & $\surd$ & $\surd$ \\
    MetaDrive \cite{li2022metadrive} & $\surd$ & * & * & - & $\surd$ & $\surd$ & $\surd$ \\
    \rowcolor{light-gray} NuPlan \cite{caesar2021nuplan} & - & - & - & - & $\surd$ & - & $\surd$ \\
    InterSim \cite{sun2022intersim} & - & - & - & $\surd$ & $\surd$ & $\surd$ & $\surd$ \\
    \rowcolor{light-gray} TBSim \cite{xu2023bits} & - & - & - & $\surd$ & $\surd$ & $\surd$ & $\surd$ \\
    LimSim \cite{wenl2023limsim} & - & $\surd$ & * & - & $\surd$ & - & $\surd$ \\
    \rowcolor{light-gray} Waymax \cite{gulino2023waymax} & $\surd$ & - & - & - & $\surd$ & $\surd$ & $\surd$ \\
    \midrule
    Tactics2D & $\surd$ & $\surd$ & $\surd$ & $\surd$ & $\surd$ & $\surd$ & $\surd$ \\
    \bottomrule[2pt]
    \end{tabular}
\end{table*}

\subsection{Built-in environment}

Nearly half of the simulators lack a built-in training and evaluating environment for driving decision models. Consequently, users of these simulators must manually define terminal conditions and determine which elements of the traffic scenario should be reset in what manner, undermining the establishment of a widely recognized benchmark for assessment. In contrast, Tactics2D offers multiple pre-built environments, streamlining the development of driving decision-making algorithms with immediate usability.

\subsection{Custom trajectory}

More than half of the simulators cannot import and replay trajectories recorded offline by users. Although MetaDrive and CARLA allow users to record running logs within the simulator, these logs often reflect control inputs from keyboards, resulting in significant differences in acceleration and steering patterns compared to real-world scenarios. The reluctance to implement trajectory customization in many simulators stems from challenges in aligning their physics models with real-world vehicle dynamics. In response to this challenge, Tactics2D introduces a flexible standard for ingesting trajectory data by recording only its physics states. 

\subsection{Custom map}

The capability to import and display maps created by users is widely desired but seldom implemented. SUMO facilitates this process by converting XODR or OSM data into its own data structure \cite{behrisch2011sumo}. LimSim achieves the functionality by calling the interface to SUMO \cite{wenl2023limsim}. On the other hand, CARLA enables users to edit XODR maps through RoadRunner or directly model objects within the UnrealEngine \cite{dosovitskiy2017carla}, while MetaDrive utilizes CARLA's method for processing XODR maps \cite{li2022metadrive}. Tactics2D develops a unique map data structure capable of simultaneously handling open map formats XODR and OSM, allowing users to import custom maps using editors JOSM and RoadRunner.

\subsection{Dataset compatibility}

\begin{table*}[t]
    \centering
    \caption{Comparison of dataset compatibility with influential actively maintained open-source simulators.}
    \label{dataset-compatibility}
    \small
    \begin{tabular}{l | c c c c c c} \toprule[2pt]
    \textbf{Simulator} & \textbf{Argoverse} \cite{wilson2023argoverse} & \textbf{DLP} \cite{shen2020parkpredict} & \textbf{INTERACTION} \cite{zhan2019interaction} & \textbf{LevelX} \cite{krajewski2018highd, bock2020ind, krajewski2020round, moers2022exid} & \textbf{NuScenes} \cite{caesar2020nuscenes} & \textbf{WOMD} \cite{ettinger2021large} \\ \midrule
    SUMO \cite{behrisch2011sumo} & - & - & - & - & - & - \\
    \rowcolor{light-gray} CarRacing \cite{carracing} & - & - & - & - & - & - \\
    CARLA \cite{dosovitskiy2017carla} & - & - & - & - & - & - \\
    \rowcolor{light-gray} CommonRoad \cite{althoff2017commonroad} & - & - & $\surd$ & $\surd$ & - & - \\
    highway-env \cite{highway-env} & - & - & - & - & - & - \\
    \rowcolor{light-gray} SMARTS \cite{zhou2020smarts} & - & - & - & - & - & - \\
    MetaDrive \cite{li2022metadrive} & - & - & - & - & - & - \\
    \rowcolor{light-gray} NuPlan \cite{caesar2021nuplan} & - & - & - & - & $\surd$ & - \\
    InterSim \cite{sun2022intersim} & - & - & - & - & $\surd$ & $\surd$ \\
    \rowcolor{light-gray} TBSim \cite{xu2023bits} & - & - & - & - & $\surd$ & $\surd$ \\
    LimSim \cite{wenl2023limsim} & - & - & - & - & - & $\surd$ \\
    \rowcolor{light-gray} Waymax \cite{gulino2023waymax} & - & - & - & - & - & $\surd$ \\
    \midrule
    Tactics2D & $\surd$ & $\surd$ & $\surd$ & $\surd$ & $\surd$ & $\surd$ \\
    \bottomrule[2pt]
    \end{tabular}
\end{table*}

An interesting observation from Table \ref{feature-comparison} is that most simulators cannot customize trajectory and map data while remaining compatible with various trajectory datasets. This limitation arises from the inherent challenge of unifying maps and trajectories recorded by different providers, often in different formats and with distinct attributes. Specifically, NuScene and WOMD tailor parsers to process their own datasets \cite{caesar2021nuplan, gulino2023waymax}. To overcome the barrier of developing algorithms across multiple datasets, Tactics2D extensively investigates popular public trajectory datasets and develops a comprehensive data structure that ensures compatibility with them. All the public trajectory datasets listed in Table \ref{dataset-compatibility} can be loaded seamlessly without loss of information.

\subsection{Interactive NPCs}

While many simulators claim to have implemented interactive NPCs, the quality of their interaction models for traffic participants varies significantly. Moreover, the interactive behavior model remains under exploration and has undergone rapid iteration in recent years. Therefore, Tactics2D not only implements modular reaction algorithms based on the theories adopted in \cite{erdmann2015sumo, sun2022intersim} but also provides an interface within the traffic participant instance for users to define and introduce their own behavior models.

\subsection{Multi-agent}

A simulation environment with multi-agent capabilities can aid researchers in observing how their methods perform when interacting with other agents following the same policy. In Tactics2D, the multi-agent simulation is facilitated by the scenario manager module. Users can configure the backend of each traffic participant within the scenario manager, allowing them to simulate cooperatively/competitively interactions involving multiple agents. Meanwhile, the traffic event detector ensures compliance with traffic regulations to maintain legal behaviors.

\subsection{Lightweight}

Like other simulators tailored for driving decision-making tasks, Tactics2D is lightweight in terms of storage space and computational resources. It can be installed by PyPI command and its fundamental functions do not require a GPU.

\section{Future Works}
\label{future}

The foundational elements of Tactics2D have been fully integrated, yet continuous efforts are underway to enrich its functionalities and capacities.

 Tactics2D aims to enhance the fidelity of NPCs' behavioral models by considering additional state factors and deriving realistic responses from trajectory datasets. This endeavor will involve both reproducing existing algorithms and devising new methods \cite{xu2023bits, wenl2023limsim}. This enhancement is expected to introduce a more convincing simulation.

Since Tactics2D primarily focuses on furnishing traffic scenarios for training and testing decision-making models, it simplifies sensory data and vehicle models. To enable the evaluation of a comprehensive autonomous driving system, Tactics2D is set to collaborate with the under-development 3D simulator, Tactics. The co-simulation will integrate interactive NPCs and generative traffic scenarios from Tactics2D with the realistic sensor and physics simulation from Tactics.

In response to community suggestions, Tactics2D plans to incorporate built-in interfaces for CARLA and SUMO. These interfaces will enable seamless import/export of maps and access to NPCs' behavior models, promoting versatility and utility across various applications.

 Tactics2D is a simulation project under long-term maintenance to foster an active community for discussion. Proposed functionalities on the GitHub Issue page and Discord Channel will be thoroughly evaluated and continuously integrated into the simulator's development roadmap.

\section{Online Supporting Materials}

\subsection{Insight of public trajectory datasets}

In the `trajectory\_data\_analysis\ directory of \url{https://github.com/SCP-CN-001/trajectory_dataset_support/tree/main}, a comprehensive analysis of various public trajectory datasets \cite{krajewski2018highd, bock2020ind, krajewski2020round, moers2022exid, wilson2023argoverse, shen2020parkpredict, zhan2019interaction, caesar2020nuscenes, ettinger2021large} is provided. This analysis clearly shows that the speed, mean speed, and steering angle significantly vary across different maps. Moreover, as the traffic scenarios are collected from diverse cities, the deviation in these metrics further amplifies. Furthermore, the proportion of different types of traffic participants varies significantly across different traffic scenarios. This disparity underscores the importance of utilizing real-world trajectory and map data to capture the intricacies of driving scenarios. It also emphasizes the ongoing need for developing high-fidelity scenario generators to advance driving decision-making systems.

\subsection{Demonstration of available scenarios}

The current version of Tactics2D supports various traffic scenarios, including the highway, intersection, roundabout, parking lot, and racing track environments. Visualizations of these scenarios are available at \url{https://tactics2d.readthedocs.io/en/latest/dataset-support/}. For guidance on training and evaluating models in these environments, refer to the `Tutorial' section at \url{https://tactics2d.readthedocs.io/en/latest/}.


\bibliographystyle{IEEEtran}
\bibliography{main}

\end{document}